\documentclass[10pt,twocolumn,letterpaper]{article}

\usepackage[pagenumbers]{cvpr} 

\usepackage[dvipsnames]{xcolor}

\definecolor{myPurple}{rgb}{0.4, .0, .8}
\definecolor{myGreen}{rgb}{0, .8, .3}
\definecolor{myRed}{rgb}{0.8, .2, .2}
\definecolor{myOrange}{rgb}{0.7, 0.45, 0.2}
\definecolor{myBlue}{rgb}{.0, .0, 1.0}
\definecolor{myBlue2}{rgb}{.0, .0, 0.5}
\definecolor{myBlack}{rgb}{.0, .0, 0.0}

\definecolor{cvprblue}{rgb}{0.21,0.49,0.74}
\usepackage[pagebackref,breaklinks,colorlinks,citecolor=cvprblue]{hyperref}
\usepackage{comment}

\usepackage{array}
\usepackage{multirow}

\def\paperID{326} 
\def\confName{3DV\xspace}
\def\confYear{2025\xspace}

\def\eg{\emph{e.g.}}

\title{GaussianAvatar-Editor: Photorealistic Animatable Gaussian Head Avatar
Editor}

\author{
Xiangyue Liu\textsuperscript{\rm 1} \ \ \ Kunming Luo\textsuperscript{\rm 1} \ \ \ Heng Li\textsuperscript{\rm 1} \ \ \ Qi Zhang\textsuperscript{\rm 2} \ \ \ Yuan Liu\textsuperscript{\rm 1} \ \ \ Li Yi\textsuperscript{\rm 3} \ \ \ Ping Tan\textsuperscript{\rm 1}\footnotemark[2] \\
\\
\textsuperscript{\rm 1} Hong Kong University of Science and Technology \\ 
\textsuperscript{\rm 2} Tencent AI Lab \ \ \
\textsuperscript{\rm 3} Tsinghua University 
}

\begin{document}

\twocolumn[{%
\renewcommand\twocolumn[1][]{#1}%
\maketitle
\begin{center}
    \centering
    \captionsetup{type=figure}
    \includegraphics[width=0.85\linewidth]{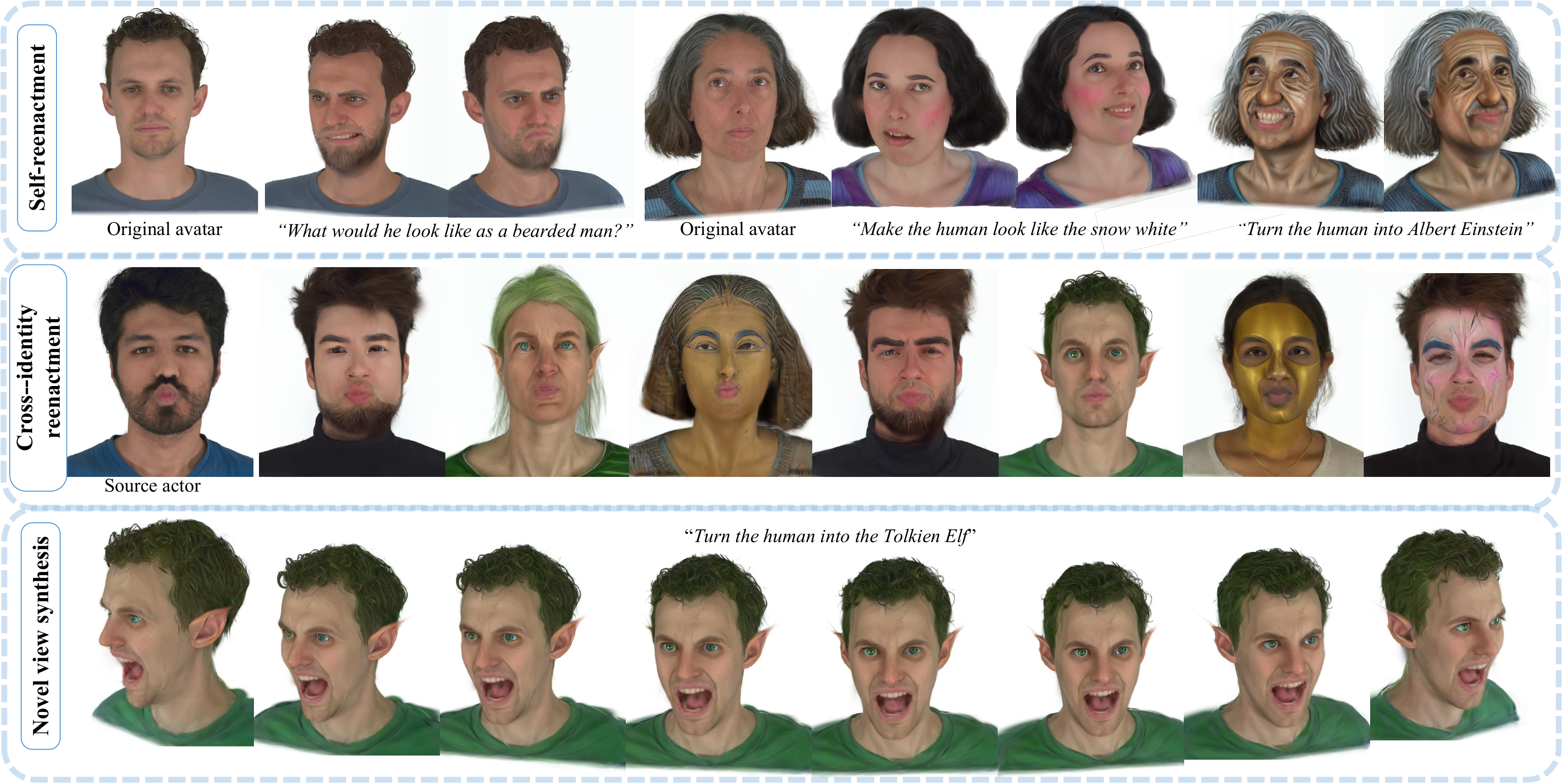}
    \captionof{figure}{
    We introduce GaussianAvatar-Editor, a method for text-driven editing of animatable Gaussian head avatars with fully controllable expression, pose, and viewpoint. We show qualitative results of our GaussianAvatar-Editor at the inference time above. Our edited avatars can achieve photorealistic editing results with strong spatial and temporal consistency.
    }
    \label{fig:teaser}
\end{center}%
}]



\begin{abstract}
We introduce GaussianAvatar-Editor, an innovative framework for text-driven editing of animatable Gaussian head avatars that can be fully controlled in expression, pose, and viewpoint. Unlike static 3D Gaussian editing, editing animatable 4D Gaussian avatars presents challenges related to motion occlusion and spatial-temporal inconsistency. To address these issues, we propose the Weighted Alpha Blending Equation (WABE). This function enhances the blending weight of visible Gaussians while suppressing the influence on non-visible Gaussians, effectively handling motion occlusion during editing. Furthermore, to improve editing quality and ensure 4D consistency, we incorporate conditional adversarial learning into the editing process. This strategy helps to refine the edited results and maintain consistency throughout the animation. By integrating these methods, our GaussianAvatar-Editor achieves photorealistic and consistent results in animatable 4D Gaussian editing. We conduct comprehensive experiments across various subjects to validate the effectiveness of our proposed techniques, which demonstrates the superiority of our approach over existing methods. More results and code are available at: \url{https://xiangyueliu.github.io/GaussianAvatar-Editor/}.


\end{abstract}    
\section{Introduction}
\label{sec:intro}

The 3D reconstruction of head avatars using the radiance field-based representation \cite{mildenhall2020nerf} has shown unparalleled photorealistic rendering quality and impressive animatable results. This is critical for visual communications, immersive telepresence, movie production, and augmented or virtual reality. Recently, 3D Gaussian Splatting (3DGS)~\cite{kerbl20233d} proposes a GPU-friendly differentiable rasterization pipeline that employs an explicit point-based representation, achieving superior rendering quality compared to NeRF for novel view synthesis while maintaining real-time performance. 3DGS has been utilized in various downstream applications, particularly head avatar reconstruction \cite{qian2023gaussianavatars,saito2024rgca,wang2024gaussianhead,xu2023gaussianheadavatar,chen2023monogaussianavatar} with real-time rendering for novel poses and expressions.

Although 3DGS-based avatar reconstruction exhibits remarkable animations, it is essential to incorporate advanced customization options, such as texture editing, shape manipulation, and accessory generation, to accommodate the diverse needs of users. With the rapid advancement of 2D diffusion-like text-to-image (T2I) techniques \cite{rombach2022high, saharia2022photorealistic}, generative text-driven 3D editing \cite{haque2023instruct, zhuang2023dreameditor, zhuang2024tip} has emerged as a novel approach, complementing previous 3D style transformation and shape manipulation methods \cite{perez2024styleavatar, wang2023styleavatar, chan2022efficient}. Specifically, Instruct-NeRF2NeRF \cite{haque2023instruct} employs an image-based diffusion model to modify the rendered image by the text prompt, and subsequently updates the 3D radiance field with the edited image. Text-driven 3D editing framework produces promising results on view consistency, enabling more flexible and enhanced editing through text control. 

To enable the editing of head avatars, a straightforward solution is to introduce such text-driven editing strategy in Gaussian head avatars. However, challenges remain in editing animatable 3D head avatars using text instructions, particularly regarding anti-occlusion editing in motion-occlusion regions (\textit{e.g.}, teeth occluded by the mouth, eyeballs occluded by eyelids, nosehole occluded by the nose tip) and maintaining spatial-temporal consistency in the editing region throughout the animation process as shown in Fig.~\ref{fig:fig3} and Fig.~\ref{fig:ours_ab1}. Specifically, motion occlusions occur when certain parts of the avatar are temporarily obscured by other parts, such as when the lips obscure the teeth as shown in Fig.~\ref{fig:ours_ab1}. The occluders can easily affect the Gaussians of the occluded part, leading to artifacts and inconsistencies when animating edited avatars. Meanwhile, the edited images at different timesteps and viewpoints may not be consistent with each other, which also greatly degenerates the generation quality.

To address the above challenges, we introduce our method, \textit{GaussianAvatar-Editor}, to edit animatable head avatars. 
Specifically, to overcome the incorrect editing caused by occlusions, we propose a novel activation function applied in Gaussian alpha blending for anti-occlusion. To improve the 4D consistency, we apply adversarial learning in the editing framework to reduce the impact of inconsistent supervision signals from diffusion-based editors, greatly improving editing quality. 
Some results from our GaussianAvatar-Editor in several challenging scenarios are shown in Fig.\ref{fig:teaser}. In both qualitative and quantitative comparisons, our method consistently outperforms existing methods in novel views, poses, and expressions.


To summarize, our main contributions are threefold.
\begin{enumerate}
\setlength{\itemsep}{0pt}
\setlength{\parsep}{0pt}
\setlength{\parskip}{0pt}
    \item[-] We propose an innovative activation function applied in the Gaussian alpha blending, making our framework robust to multi-layer surfaces.
    \item[-] We introduce an adversarial learning framework to learn from the 2D diffusion-based editor, which reduces the impact of inconsistent supervision signals and improves the quality of animatable head editing.
    \item[-] Building on the proposed activation function and adversarial learning, we introduce GaussianAvatar-Editor, which achieves high-quality editing and ensures spatio-temporal consistency in challenging scenarios.
\end{enumerate}
\begin{figure*}
\begin{center}
   \includegraphics[width=0.92\linewidth]{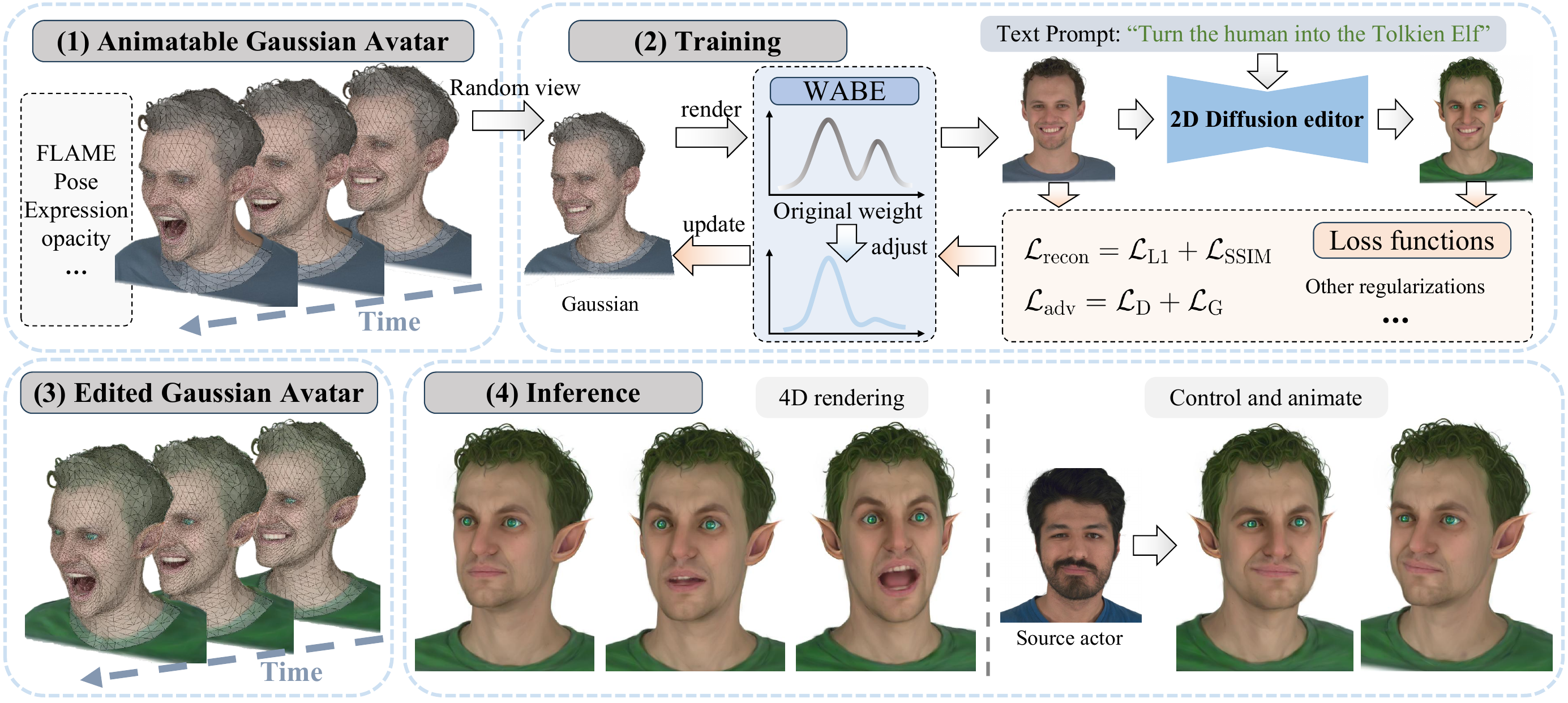}
\end{center}
\vspace{-15pt}
   \caption{The overview of our method. We follow a render-edit-aggregate optimization pipeline as in Instruct-NeRF2NeRF~\citep{haque2023instruct}. We introduce a Weighted Alpha Blending Equation (WABE) to overcome the motion occlusion problem and our novel loss functions to enhance the spatial-temporal consistency. Our edited avatars can generate high-quality and consistent 4D renderings and can be controlled by other actors.}   
\label{fig:fig2}
\end{figure*}

\section{Related Works}

\noindent\textbf{Text-driven Editing.}
Diffusion models for text-to-image generations \cite{rombach2022high, saharia2022photorealistic} have impressive capability in generating diverse, high-quality images from textual prompts. This innovation has led to a variety of applications, such as image-to-image translation \cite{meng2021sdedit,bar2022text2live,hertz2022prompt,brooks2023instructpix2pix,mokady2023null,kawar2023imagic} and controllable generation \cite{voynov2023sketch,zhang2023adding}. The advancements have also brought significant progress to numerous 3D tasks, such as text-driven editing, benefiting from the abundant prior knowledge of pre-trained text-to-image models. Some works \cite{wang2022clip,bao2023sine,song2023blending,wang2023nerf} leverage a CLIP model to edit reference images and lift to 3D space through NeRF optimization. Instruct-NeRF2NeRF \cite{haque2023instruct} and Instruct 3D-to-3D \cite{kamata2023instruct} distill 3D scenes from a pretrained text-driven image editing model\cite{brooks2023instructpix2pix}. TextDeformer \cite{gao2023textdeformer} and Texture \cite{richardson2023texture} achieve geometry and texture modification according to text prompts, respectively.  Vox-E \cite{sella2023vox} and DreamEditor \cite{zhuang2023dreameditor} leverage the SDS loss \cite{poole2022dreamfusion} to perform local editing in 3D space. TIP-Editor \cite{zhuang2024tip} introduces a novel approach for accurately controlling the appearance of specified 3D regions with both text and image prompts. The editing ability is further enhanced by upscaling to 4D with dynamic scene representations like 4D NeRF. Control4D \cite{shao2023control4d} combines 4D representation with GAN to achieve better spatial-temporal consistency in dynamic scene editing. Our method capitalizes on 3DGS, which achieves real-time renderings with high-quality and text-driven editing ability of InstructPix2Pix \cite{brooks2023instructpix2pix}, and achieves spatial-temporal consistent editing with given textual instructions.

\noindent\textbf{Head Avatar Reconstruction and Editing.}
A main line of head avatar reconstruction integrates human priors with neural representations. For instance, NerFACE \cite{gafni2021dynamic} conditions a dynamic NeRF on extra facial expression parameters from the 3DMM model, to reconstruct the 4D facial avatar from monocular video. IMAvatar \cite{zheng2022avatar} represents the expression- and pose-related deformations from the canonical space via learned blendshapes and skinning fields, allowing generalization to unseen poses and expressions. INSTA \cite{zielonka2023instant} reconstructs a deformable radiance field based on neural graphics primitives and greatly accelerates the training and inference. Recently, many 3DGS-based methods \cite{qian2023gaussianavatars,dhamo2023headgas,zhao2024psavatar,saito2023relightable} have shown superior performance in speed and texture. AvatarStudio \cite{pan2023avatarstudio} reconstructs dynamic digital avatars from multi-view videos and achieves editing by applying a text-driven diffusion model individually on multiple keyframes and optimizing to a unified appearance volume. Thus, its editing cannot be generalized to new expressions and may result in artifacts while handling expressions with significant changes. We address these challenges by considering the differences arising from expressions and poses and achieving high-quality editing that maintains spatial-temporal and spatial-animatable consistency.

\noindent\textbf{3D Gaussian Head Avatar.}
Various methods \cite{qian2023gaussianavatars,saito2024rgca,wang2024gaussianhead,xu2023gaussianheadavatar,chen2023monogaussianavatar} attempt to bring Gaussian Splatting to dynamic 3D human head avatar reconstruction. GaussianAvatars~\citep{qian2023gaussianavatars} proposes binding 3D Gaussian to the FLAME~\citep{li2017learning} model mesh. Specifically, GaussianAvatars~\citep{qian2023gaussianavatars} initializes a 3D
Gaussian at the center of each FLAME~\citep{li2017learning} model triangle and uses a binding strategy to support Gaussian splats that densify and prune while maintaining the binding relations. Then, it optimizes the 3D Gaussian and the FLAME~\citep{li2017learning} model in an end-to-end fashion. In this work, we pioneer the adaptation of 3D Gaussian splatting to Animatable Head Avatar Editing tasks, aiming to achieve photorealistic editing and reenactment to different actors, executing the advantages of Gaussian Splitting representation for the first time in this context. Considering gradients from visible pixels (non-occluded regions) may erroneously propagate to non-visible Gaussians (occluded parts), we specifically designed an activation function for Gaussian alpha blending to handle the motion-occluded regions.


\section{Preliminary} \label{sec:preliminaries}

\subsection{3D Gaussian Splatting}
Gaussian Splatting~\citep{kerbl20233d} represents 3D scenes using Gaussian spheres $\{G_k\ |\ k = 1,\dots,K\}$, where each Gaussian $G_k$ is defined by the point center $\mathbf{\mu_k}$, and a covariance matrix $\mathbf{\Sigma_k}$ as,
\begin{equation}
G_k(\mathbf{x}) = e^{-\frac{1}{2} (\mathbf{x}-\mathbf{\mu_k})^\top \mathbf{\Sigma_k}^{-1} (\mathbf{x}-\mathbf{\mu_k})}.
\end{equation}
The covariance matrix $\mathbf{\Sigma_k}$ is parameterized by a rotation matrix $\mathbf{R_k}$ and a scaling matrix $\mathbf{S_k}$ as $\mathbf{\Sigma_k} = \mathbf{R_kS_kS_k^{\top}R_k^{\top}}$.


During rendering, 3DGS~\citep{kerbl20233d} employs spherical harmonics $\mathbf{c_k}$ to model view-dependent color and applies $\alpha$-blending of different Gaussians according to the depth order $1,...,K$ as,
\begin{equation}
\mathbf{C(x)} = \sum_{k=1}^{K} \mathbf{c_k}\alpha_k \prod_{j=1}^{k-1} (1-\alpha_j).
\label{eq:render}
\end{equation}

\subsection{Gaussian Avatar}
The work of GaussianAvatars~\citep{qian2023gaussianavatars} binds 3D Gaussians $\mathcal{G}$ to the underlying animatable FLAME~\citep{li2017learning} model to represent a head avatar $M$ by
\begin{equation}
    M(\beta,\theta,\psi,\mathcal{G})=W(T(\beta,\theta,\psi),J(\beta),\theta,\mathcal{G}),
\end{equation}
where $\beta$ is the shape, $\theta$ is the pose, $\psi$ is the expression, $W$ means deformation with predefined skinning weights, $T$ is the pose-dependent shap, and $J$ is the joint points. Gaussians $\mathcal{G}$ are defined on the FLAME model and will be transformed along with the motion of the head avatar. It achieves photorealistic rendering and controllable animation at the same time. In this paper, we design a text-driven method to edit Gaussian avatars.

\begin{figure}
\begin{center}
   \includegraphics[width=0.82\linewidth]{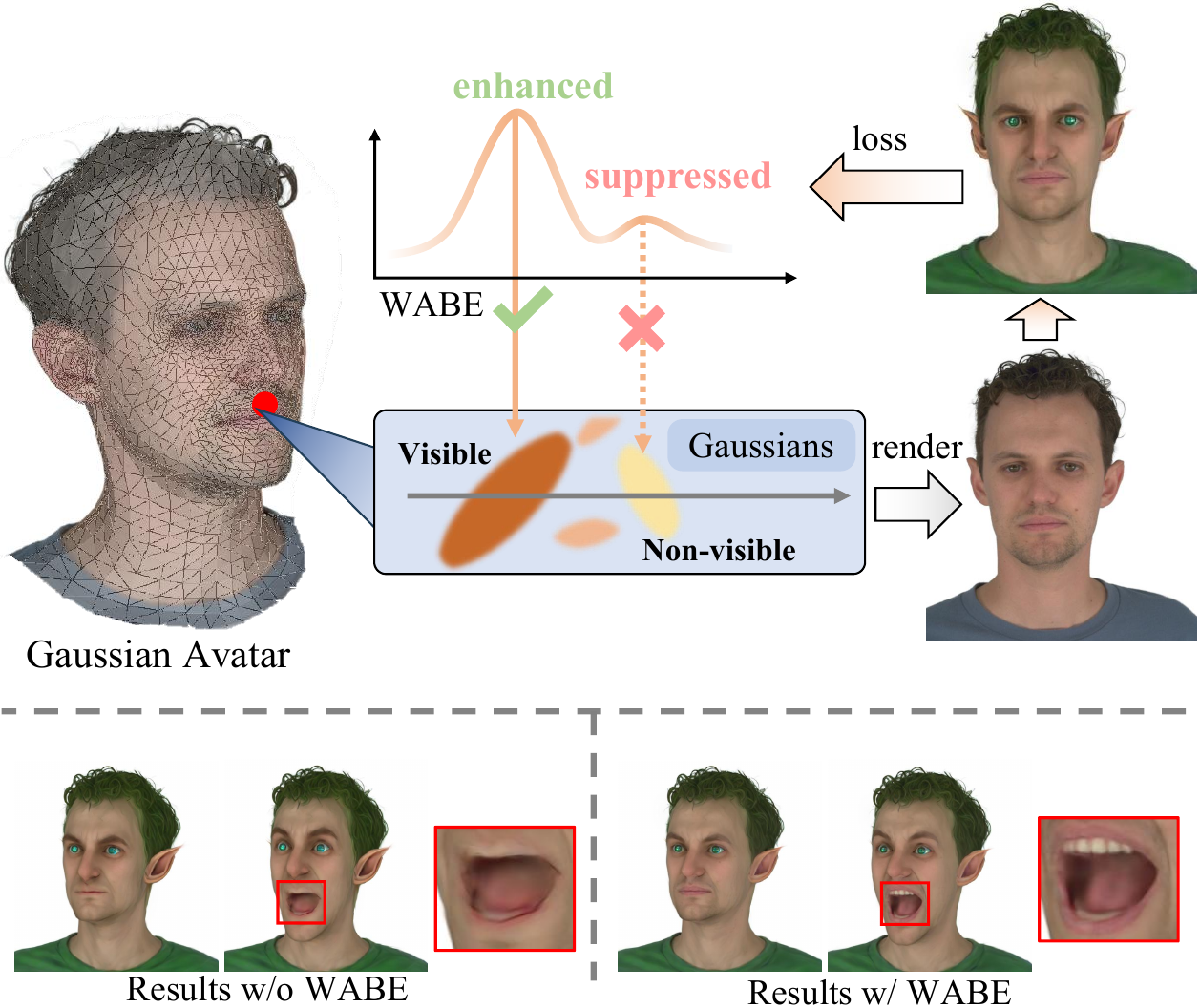}
\end{center}
\vspace{-15pt}
   \caption{
   Illustration of the Weighted alpha blending equation (WABE), which is adjusted to suppress non-visible parts while enhancing visible parts. Lower left: results when WABE is disabled. Lower right: when WABE is enabled, motion-occluded regions like teeth and tongue can be successfully optimized. 
   }   
\label{fig:fig3}
\end{figure}

\section{Method}
The overview of our proposed method is illustrated in Fig.~\ref{fig:fig2}. Given an animatable Gaussian avatar built according to the method in GaussianAvatars~\citep{qian2023gaussianavatars}, we follow the render-edit-aggregate method similar to Instruct-NeRF2NeRF~\citep{haque2023instruct} to update the avatar gradually. Specifically, we first randomly sample a training view and render an image using the Gaussian avatar. We then edit this rendered image with 2D diffusion-based editors~\cite{brooks2023instructpix2pix} according to the text prompt provided by users. Finally, we compute loss functions between the rendered and edited images and back-propagate the gradients to refine the Gaussian avatar. 

\subsection{Challenges in Gaussian Avatar Editing}
However, unlike reconstructing Gaussian avatars from multi-view videos, text-driven editing of these Gaussian avatars presents significant challenges. 

{\bf Motion occlusion.}
A key challenge is brought by the occlusions in the motion sequence, which complicates the convergence of the optimization process. Specifically, when optimizing 3D Gaussians using gradients derived from supervision images, the $\alpha$-blending rendering technique in 3DGS~\citep{kerbl20233d} updates all 3D Gaussians indiscriminately, despite whether these Gaussians are visible from the current viewpoint. 
In our scenario, gradients from visible pixels (e.g., pixels on occluders) may erroneously be propagated to invisible Gaussians (e.g., pixels on occluded parts). For example, as shown in Fig.~\ref{fig:ours_ab1}, the lip might occlude the teeth, the eyelid might occlude the eyeball, the nose might occlude the nostril, etc. When occlusion happens, the gradient should be stopped at occluders without affecting the occluded parts. 

{\bf 4D consistency.} 
Another key challenge is maintaining 4D spatial and temporal consistency after editing, \eg, the same facial point should be the same over time and across views after editing. While some recent works, such as Instruct-NeRF2NeRF~\citep{haque2023instruct}, introduce a render-edit-aggregate method to mitigate multi-view inconsistency in static 3D scene editing, ensuring 4D consistency is significantly more challenging.

\subsection{Occlusion-aware Rendering and Editing}
\label{sec:occlusion-aware}
As discussed earlier, the $\alpha$-blending in 3DGS~\citep{kerbl20233d} updates all 3D Gaussians along the ray during the training process, leading to poor editing results in regions with severe motion occlusions, as shown in Fig.~\ref{fig:fig3} and Fig.~\ref{fig:ours_ab1}. Ideally, the correct approach would be to update only the visible 3D Gaussians during the editing process while preserving the 3D Gaussians in the invisible regions. Motivated by this, we propose a modified rendering equation, referred to as the weighted alpha blending equation (WABE), specifically tailored for Gaussian avatar editing.

\begin{figure*}
\begin{center}
   \includegraphics[width=0.85\linewidth]{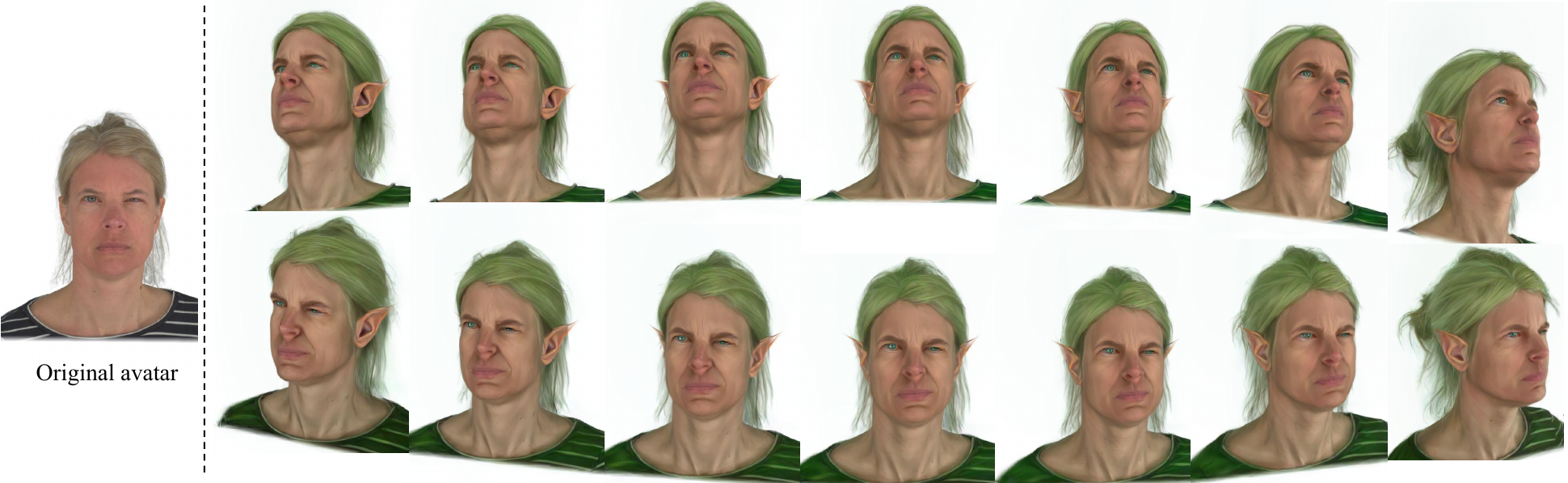}
\end{center}
\vspace{-15pt}
   \caption{Our results on novel view synthesis. We show our edited results using the text prompt \textit{“Turn her into the Tolkien Elf”}.
   }
\label{fig:ours_nv}
\end{figure*}

\begin{figure*}[t]
\begin{center}
   \includegraphics[width=0.90\linewidth]{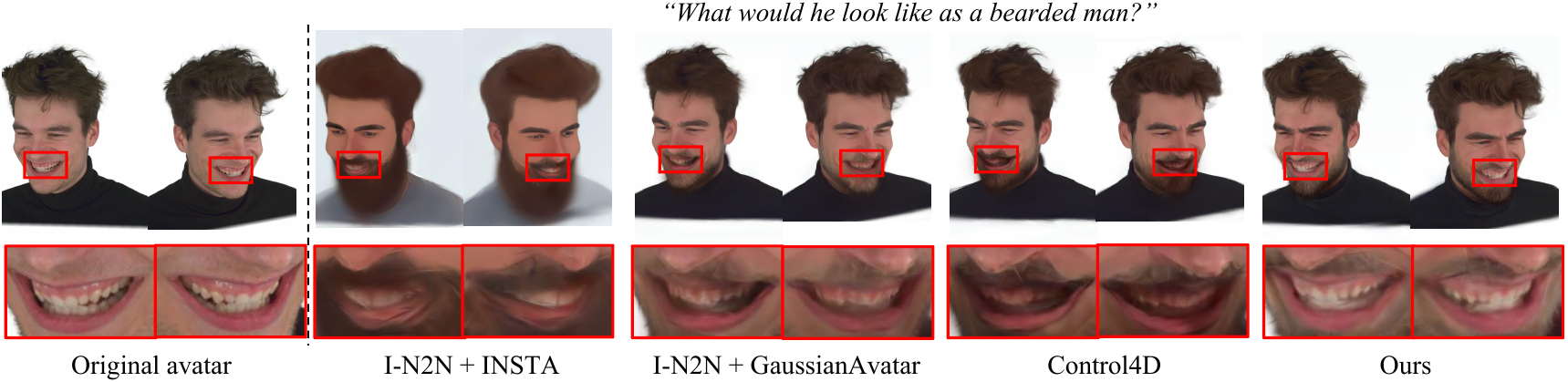}
\end{center}
\vspace{-15pt}
   \caption{Comparison on novel view synthesis. Our method produces more high-quality and multi-view consistent results than baselines.}
\label{fig:exp_1}
\end{figure*}

\textbf{Weighted alpha blending equation (WABE).} 
Ideally, we only want to update the visible Gaussian during editing while keeping the invisible Gaussian unchanged. This inspires us to seek a blending function that can make the editing process aware of the visible and invisible parts. We replace the original $\alpha$-blending function, i.e. Eq.~\ref{eq:render}, of the Gaussian splatting as follows,
\begin{equation}
\mathbf{C(x)} = \sum_{k=1}^{K} w_k\mathbf{c}_k\alpha_k \prod_{j=1}^{k-1} (1-\alpha_j),
\label{eq:adj_render}
\end{equation}
where we add an additional term $w_k$ here to model the visibility of the current Gaussian. Invisible Gaussians will have zero weights.  
To achieve this goal, as illustrated in Fig.~\ref{fig:fig3}, we design the weighted function $w_k$ as follows,
\begin{equation}
w_k = e^{-\beta (1-\prod_{j=1}^{k-1} (1-\alpha_j)))},
\label{eq:adj}
\end{equation}
where $\beta$ controls the distribution of weights between layers. $\prod_{j=1}^{k-1} (1-\alpha_j)$ is the probability of not being occluded by Gaussians in front of the current one, i.e. the visibility of the current Gaussian. According to our definition, $w_k$ will decrease to 0 with the reduction of visibility, and  $w_k$ equals 1 if the Gaussian is fully visible.
A larger value of $\beta$ makes the weights $w_k$ change more fast, leading to more pronounced transitions between layers. This produces a stronger blending effect with sharper changes in transparency. We set $\beta$ to 6 in all of our experiments. As shown in Fig.~\ref{fig:fig3}, when our WABE is enabled, the editing `Turn him into the Tolkien Elf' only affects the skin and does not change the teeth. 
\subsection{4D Consistent Editing}

\noindent{\bf Editing reconstruction loss.}
We apply the InstructPix2Pix~\cite{brooks2023instructpix2pix} model to generate edited images $\mathbf{E}_i^t$. Then, to edit the Gaussian avatar, we use the reconstruction loss as the L1 norm and SSIM loss~\citep{johnson2016perceptual} between the rendered image $\mathbf{C}_i^t$ and the edited image $\mathbf{E}_i^t$ as follows,
 \begin{equation}
\begin{aligned}
\mathcal{L}_{\textrm{recon}} &= \mathcal{L}_{\textrm{L1}} + \mathcal{L}_{\textrm{SSIM}} \\ 
&= \left \| \mathbf{C}_i^t - \mathbf{E}_i^t\right \|_{1} + SSIM(\mathbf{C}_i^t - \mathbf{E}_i^t),
\end{aligned}
\end{equation}
where $i$ is the viewpoint index, and $t$ is the time index. 

Due to the lack of temporal and spatial consistency in the images edited by instructions, the supervision in Gaussian splatting optimization might lead to conflicts. Inspired by Instruct-NeRF2NeRF~\citep{haque2023instruct}, we extend its render-edit-aggregate pipeline to the 4D space to gradually optimize the origin avatar towards the final convergent result. Specifically, we first randomly sample a training view $i$ and a time $t$ and render an image $\mathbf{C}_i^t$ using the Gaussian avatar. We then edit this rendered image with 2D diffusion-based editors~\cite{brooks2023instructpix2pix} according to the text prompt provided by users. Finally, we compute loss functions between the rendered image $\mathbf{C}_i^t$ and the edited image $\mathbf{E}_i^t$ and back-propagate the gradients to refine the Gaussian avatar. 




\begin{figure*}[h]
\begin{center}
  \centering
    \includegraphics[width=0.90\linewidth]{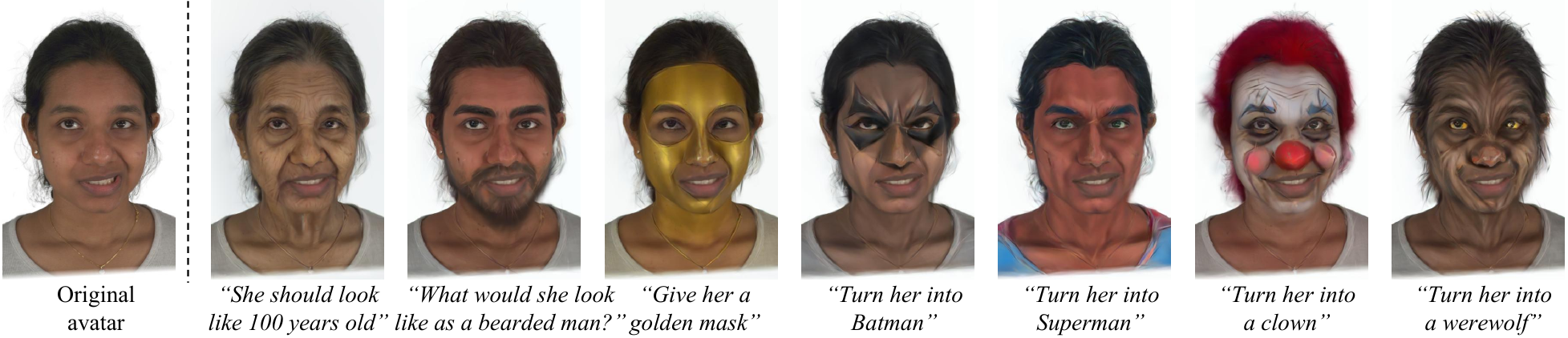}
\end{center}
\vspace{-15pt}
   \caption{Our results on self-reenactment. Self-reenactment renders held-out unseen head pose and expressions from 16 training camera viewpoints. The bottom part shows the text prompts. 
   }
\label{fig:ours_self}
\end{figure*}

\begin{figure*}[h]
\begin{center}
  \centering
    \includegraphics[width=0.90\linewidth]{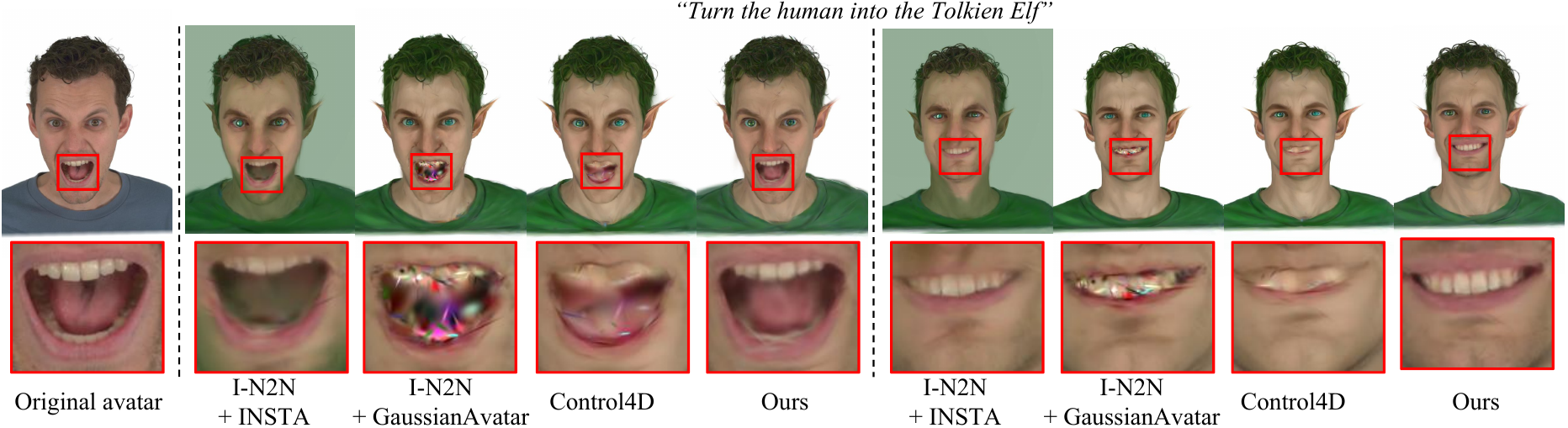}
\end{center}
\vspace{-15pt}
   \caption{Comparison of self-reenactment. Our edited avatar can correctly produce detailed facial features under unseen expressions and head poses from the same subject. }
\label{fig:exp_2}
\end{figure*}

{\bf Temporal adversarial learning.} 
Since the temporal consistency of the instruction-edited images is not ensured, relying solely on reconstruction loss like Instruct-NeRF2NeRF~\citep{haque2023instruct} often leads to blurry or distorted artifacts in results, especially in animations. Thus, we introduce a temporal adversarial learning scheme to improve consistency in different time steps.

Previous work~\cite{liu2024genn2n} has demonstrated the effectiveness of conditional adversarial training in preventing blurry rendered images by training a discriminator to determine true or fake images of different viewpoints. 
This inspires us to extend this conditional adversarial loss to enforce temporal consistency, which alleviates blurry artifacts in rendered images. 
More specifically, we train a discriminator $\mathcal{D}$ 
to distinguish real and fake image pairs. The real image pairs $\mathit{P}_{real}$ 
consists of ${\mathbf{E}_i^t}$ and ${\mathbf{E}_i^t-\mathbf{E}_i^k}$
where ${\mathbf{E}_i}$ is the edited image from the 2D image editor InstructPix2Pix~\cite{brooks2023instructpix2pix}, and $t$, $k$ means adjacent timestep. Similarly, a fake pair $\mathit{P}_{fake}$ consists of the rendered images ${\mathbf{C}_i^t}$ and ${\mathbf{C}_i^t-\mathbf{E}_i^k}$. 
The pairs are concatenated in RGB channels and fed into the discriminator $\mathcal{D}$. We optimize the discriminator $\mathcal{D}$ and the edited Gaussian avatar with the following objective functions,
\begin{equation}\label{eq:adversarial}
\begin{aligned}
 \mathcal{L}_{\textrm{D}} &= \mathbb{E}_{\mathbf{R}}[-log(\mathcal{D}(\mathit{P}_{real}))]+\mathbb{E}_{\mathbf{F}}[-log(1-\mathcal{D}(\mathit{P}_{fake}))], \\ 
\mathcal{L}_{\textrm{G}} &= \mathbb{E}_{\mathbf{F}}[-log(\mathcal{D}(\mathit{P}_{fake}))]. \\
\end{aligned}
\end{equation}
In this adversarial loss, we compare not only the edited images with the rendered images but also the differences on different timesteps, which forces the model to learn the temporal consistency for better editing quality.


\subsection{Optimization and Regularization}
During the training process, we jointly optimize the loss functions mentioned above: $\mathcal{L}_{\textrm{recon}}$ and $\mathcal{L}_{\textrm{G}}$ for the edited Gaussian splattings, and $\mathcal{L}_{\textrm{D}}$ for the discriminator. Inspired by GaussianAvatars~\citep{qian2023gaussianavatars}, we also regularize the Gaussian's position and scales to make Guassians close to the underlying FLAME~\citep{li2017learning} model by $\mathcal{L}_{\textrm{const}}$. The total loss formula is expressed as follows: 
\begin{equation}
\mathcal{L} = \lambda_1 \mathcal{L}_{\textrm{recon}}  + 
\lambda_2 \mathcal{L}_{\textrm{D}} + 
\lambda_3 \mathcal{L}_{\textrm{G}} + 
\lambda_4 \mathcal{L}_{\textrm{const}},
\end{equation}
where $\lambda$ means the weights for the loss and we set $\lambda_1=10$, $\lambda_2=0.01$, $\lambda_3=0.01$, and $\lambda_4=10$ in all of our experiments. 
The weights can be adjusted to prioritize different aspects of the training objective, such as reconstruction accuracy, adversarial training, and the perceptual quality. 

\subsection{Inference}\label{method: inference}
After the optimization of our method, the edited Gaussian head avatar can render the target novel views conditioned on the given expression and pose parameters. 


\begin{figure*}
\begin{center}
  \centering
    \includegraphics[width=0.90\linewidth]{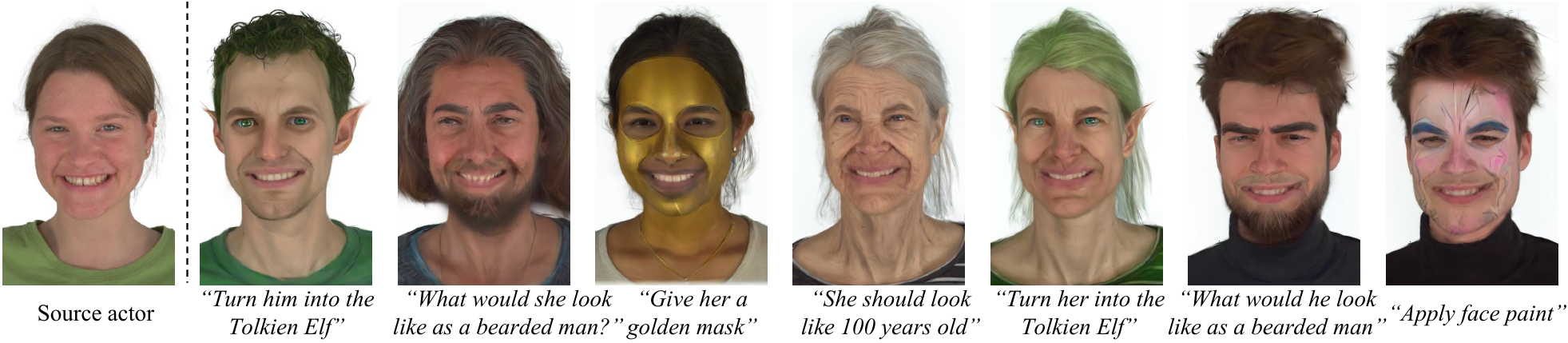}
\end{center}
\vspace{-15pt}
   \caption{Our results on cross-identity reenactment. Cross-identity reenactment animates the avatar to render images with unseen head poses and expressions from sequences of a different actor. The bottom part shows the text prompts.  
   }
\label{fig:ours_cross}
\end{figure*}

\begin{figure*}
\begin{center}
  \centering
    \includegraphics[width=0.85\linewidth]{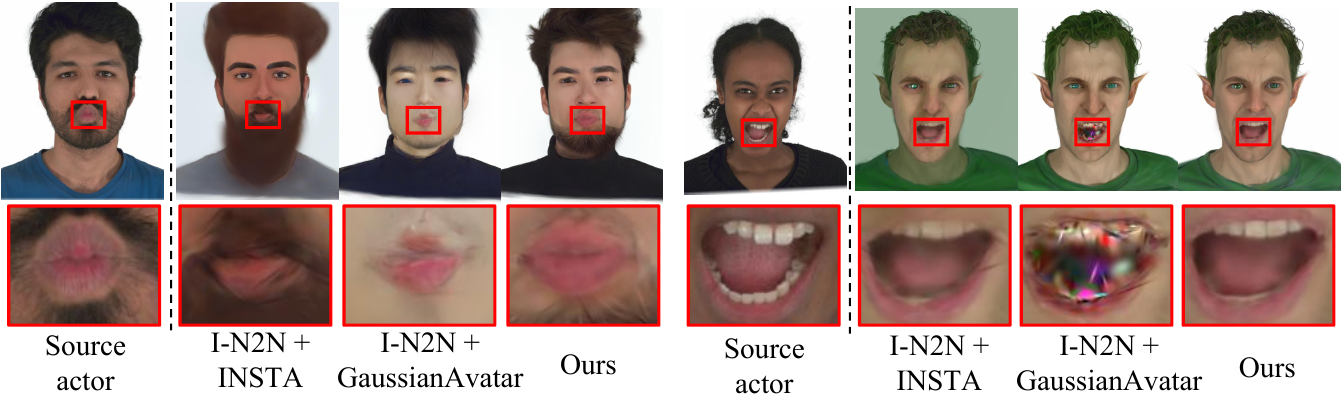}
\end{center}
\vspace{-15pt}
   \caption{Comparison of Cross-identity reenactment. Different edited avatars are controlled by the same source actor. Our method can render high-quality results with novel expressions, while baseline methods suffer from artifacts. }
\label{fig:exp_3}
\end{figure*}

\section{Experiments}
\label{sec:experiments}

\subsection{Setup}

{\bf Implementation details.} In our pipeline, we first use GaussianAvatars~\citep{qian2023gaussianavatars} to reconstruct the original animatable Gaussian head avatar from input videos. Then, we edit the original avatar using the input text prompt. Our model is trained using the Adam optimizer with a learning rate of $1e-2$, running for $10,00$ iterations per editing. The total training phase takes about $15$ minutes on one 42GB NVIDIA A100 GPU. Given a set of novel expressions, poses, and viewpoints during the inference phase, we can directly drive the edited avatar to the new pose render images of the edited Gaussian avatar. 

{\bf Dataset.} 
We conducted experiments on the NeRSemble dataset~\citep{kirschstein2023nersemble}, which consists of multi-view videos capturing the front and side views of 8 individuals from 16 camera viewpoints. There are 11 video sequences for each subject, and each video sequence contains approximately 150 frames of different expressions and movements.
The first 10 sequences include instructed facial expressions and emotions, while the last sequence records free expressions. During our experiments, all video images are downscaled to a resolution of $802\times550$.
For quantitative evaluation, we use 9 out of 10 video sequences and 15 out of 16 camera views for training and use the last video sequence (free performance) to evaluate the ability of visually cross-identity reenactment.

{\bf Evaluation Settings.} 
We evaluate the quality of the edited head avatar from three aspects: 
(1) {\bf Novel-view rendering} that uses the edited avatars to render images with training head pose and expressions from held-out camera viewpoints; (2) {\bf Self-reenactment} that renders held-out unseen head pose and expressions from 16 training camera viewpoints; (3) {\bf Cross-identity reenactment} that uses the avatar to render images with head poses and expressions from sequences of a different subject.

{\bf Metrics.} To quantitatively evaluate the performance, we employed CLIP Text-Image Direction Similarity (CLIP-S)~\citep{haque2023instruct}, CLIP Direction Consistency (CLIP-C)~\citep{haque2023instruct} to evaluate the edited results, which measure the consistency between renderings of edited avatars and input text prompts.

{\bf Baselines.} 
Since no existing method is available to achieve animatable Gaussian avatar editing, we compare our method to the most relevant approaches. 
(1) {\bf I-N2N+GaussianAvatar}. One important baseline method is to directly apply a static 3D editing approach to 4D Gaussian avatars. Specifically, we apply Instruct-NeRF2NeRF~\citep{haque2023instruct} to perform text-drive editing on the reconstructed animatable 4D Gaussians of GaussianAvatars~\citep{qian2023gaussianavatars}. Though animatable Gaussian avatar editing can be directly achieved in this way, the edited results are far from acceptable, largely due to the motion-occlusion problem presented in Sec.~\ref{sec:occlusion-aware}. (2) {\bf I-N2N+INSTA}. To compare with animatable 4D Gaussian editing based on NeRF~\citep{mildenhall2021nerf}, we apply Instruct-NeRF2NeRF~\citep{haque2023instruct} to perform text-drive editing on the reconstructed animatable 4D NeRF of INSTA~\citep{zielonka2023instant}. (3) {\bf Control4D}. Another baseline is Control4D~\citep{shao2023control4d}, a 4D Gaussian editing method designed for GaussianPlanes (spatial triplanes and 3D Gaussian flow). Note that Control4D represents head avatars as implicit parameters, which means Control4D's result cannot be re-animated.
Since Control4D's code for dynamic editing has not been released, we reimplement it based on its static Gaussian editing version. 


\subsection{Head Avatar Editing and Animation}

Quantitative results are summarized in Table~\ref{comparision_sota}. We refer readers to the video in the supplementary for more qualitative results.

\begin{figure*}
\begin{center}
  \centering
    \includegraphics[width=0.92\linewidth]{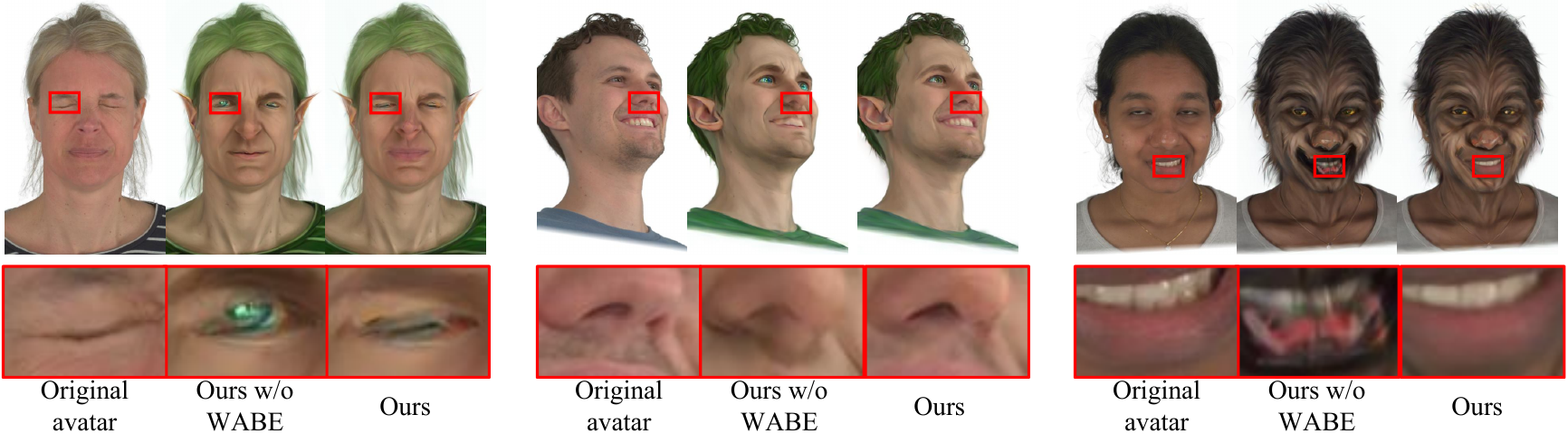}
\end{center}
\vspace{-15pt}
   \caption{Ablation study of WABE. }
\label{fig:ours_ab1}
\end{figure*}

\begin{table*}[!t]
\centering
\resizebox*{0.80 \textwidth}{!}{
\begin{tabular}{
p{2.4cm} 
>{\centering\arraybackslash}p{1.6cm}
>{\centering\arraybackslash}p{1.6cm}
>{\centering\arraybackslash}p{1.6cm}
>{\centering\arraybackslash}p{1.6cm}
>{\centering\arraybackslash}p{1.6cm}
>{\centering\arraybackslash}p{1.6cm}
}
\toprule
\multirow{2}{*}{} &\multicolumn{2}{c}{Novel view rendering} & \multicolumn{2}{c}{Self-reenactment} & \multicolumn{2}{c}{Cross-identity reenactment}\\
\cmidrule(lr){2-3} \cmidrule(lr){4-5} \cmidrule(lr){6-7}  
 & CLIP-S $\uparrow$  & CLIP-C $\uparrow$ & CLIP-S $\uparrow$  & CLIP-C $\uparrow$  & CLIP-S $\uparrow$  & CLIP-C $\uparrow$ \\
\midrule
INSTA+I-N2N & 0.181 & 0.955 & 0.042 & 0.923 & 0.043 & 0.936 \\
GA+I-N2N & 0.236 & 0.968 & 0.044 & 0.938 & 0.069 & 0.941 \\
Control4D & 0.222 & 0.980 & 0.058 & 0.938 & \slash & \slash \\
Ours w/o WABE & 0.236 & 0.968 & 0.061 & 0.948  & 0.077 & 0.950\\
Ours w/o adv & 0.266 & 0.976 & 0.077 & 0.950 & 0.070 & 0.946 \\
Ours &  \textbf{0.275} & \textbf{0.978} & \textbf{0.081} & \textbf{0.951} & \textbf{0.081} & \textbf{0.951} \\
\bottomrule
\end{tabular}
}
\vspace{-1mm}
\caption{Quantitative comparisons and ablation studies with CLIP-S and CLIP-C. We compare our method with existing methods for novel view rendering, self-reenactment, and cross-identity reenactment. 
Our method obtains superior results than other methods. 
}\vspace{-5mm}
\label{comparision_sota}
\end{table*}

{\bf Novel view rendering.} 
As shown in Fig.~\ref{fig:ours_nv}, given a Gaussian avatar and an editing prompt \textit{"Turn the human into the Tolkien Elf"}, our method can produce multi-view consistent and high-quality results.
We compare our edited avatars with existing methods for novel view synthesis. Qualitative comparison results in Fig.~\ref{fig:exp_1}, with two novel view renderings for each method. Both our method and baseline methods can produce multi-view consistent rendering results. However, the results from baseline methods are poorer, especially visible in the added beard, which is also blended on the teeth. 
In contrast, by addressing the motion occlusion problem during editing, our result can render clear and detailed teeth. Quantitative results are presented in Table~\ref{comparision_sota}, which also shows our editing results achieve better consistency with the input text than baselines.

{\bf Self-reenactment.} 
Our qualitative results are shown in Fig.~\ref{fig:ours_self}, and qualitative comparison with baselines are shown in Fig.~\ref{fig:exp_2}.
As we can see, directly applying the method Instruct-NeRF2NeRF~\citep{haque2023instruct} to INSTA~\citep{zielonka2023instant} or GaussianAvatars~\citep{qian2023gaussianavatars} results in serious artifacts at largely different head poses or facial expressions, since it is designed for editing static scenes. 
Control4D~\citep{shao2023control4d} produces better results, but its rendered images are blurry at unseen expressions. Unlike those approaches, our method obtains detailed and realistic rendering results with clear facial features even animated by unseen expressions. 
Quantity comparisons with baselines in Table~\ref{comparision_sota} also demonstrate the effectiveness of our method. 

{\bf Cross-identity reenactment.} 
To evaluate the generalization ability of our method, we further drive those edited avatars by expressions and head poses from other actors. As shown in Table~\ref{comparision_sota}, our method achieves superior CLIP-S scores and comparable CLIP-C scores as baseline methods. We also show qualitative comparison results in Fig.~\ref{fig:ours_cross} and Fig.~\ref{fig:exp_3}. As can be seen, our edited avatars can render better results than baseline methods.

\subsection{ Ablation Study}


{\bf WABE.} To validate the effectiveness of the proposed WABE for handling motion occlusion, we perform ablation experiments by disabling the WABE in our pipeline. The rendering results without WABE are also shown in Fig.~\ref{fig:fig3} and Fig.~\ref{fig:ours_ab1}. The results demonstrate that without WABE, the occluded regions like teeth when open mouth, eyeballs when closed eyes, the lips, the nosehole, etc., produce worse editing results, which leads to worse quantitative results in Table~\ref{comparision_sota}. This demonstrates the importance of WABE in handling the occlusion problem. 

{\bf Adversarial learning mechanism.} We also validate the proposed adversarial learning for spatial and temporal consistency. As shown in Table~\ref{comparision_sota}, disabling the adversarial learning loss in our pipeline decreases the test scores, especially the CLIP-C score, which demonstrates the importance of our adversarial learning mechanism. 

\section{Conclusion}

In this paper, we have presented GaussianAvatar-Editor, a text-driven framework for realistic animatable Gaussian avatar editing. For the motion occlusion problem where editing gradients would be back-propagated from non-occlusion parts to erroneously update the occlusion parts, we proposed a Weighted alpha blending equation (WABE) to replace the original Gaussian rendering function so as to suppress those erroneous updates. Moreover, to enhance the 4D supervision consistency of the editing supervision, we proposed an adversarial learning framework. By incorporating all these designs together, our method can produce high-quality, realistic editing results for 4D animatable Gaussian avatars. We have conducted comprehensive experiments on various subjects to validate the proposed methods. Both qualitative and quantitative results demonstrated that our method is superior to existing methods. 

{\bf Limitations.} GaussianAvatar-Editor utilizes FLAME model to do animation, which could not animate unmodeled parts like the tongue. We leave this for future exploration.
{
    \small
    \bibliographystyle{ieeenat_fullname}
    \bibliography{main}
}

\end{document}